\documentclass{article}
\pdfoutput=1

\usepackage[utf8]{inputenc} 
\usepackage[T1]{fontenc}    
\usepackage{hyperref}       
\usepackage{url}            
\usepackage{booktabs}       
\usepackage{amsfonts}       
\usepackage{nicefrac}       
\usepackage{microtype}      
\usepackage{lipsum}
\usepackage{graphicx}

\pdfoutput=1
\usepackage[utf8]{inputenc}
\usepackage[english]{babel}
\usepackage[T1]{fontenc}
\usepackage{amsmath}
\usepackage{hyperref}

\usepackage{tikz}
\usepackage{lipsum}
\usepackage[numbers,sort&compress]{natbib}

\usepackage{graphicx}
\usepackage{tabularx}
\usepackage{soul}
\usepackage{acronym}
\usepackage{float}
\usepackage{subcaption}
\usepackage{mathtools}
\usepackage[overload]{empheq}

\usepackage{epstopdf}
\usepackage{latexsym}
\usepackage[utf8]{inputenc}

\usepackage{tikzit}

\tikzstyle{red node}=[fill=red, draw=black, shape=circle, tikzit fill=red, tikzit draw=black]
\tikzstyle{empty_node}=[fill=white, draw=black, shape=circle, tikzit category=nodes]
\tikzstyle{blackbox}=[fill=white, draw=black, shape=rectangle]
\tikzstyle{redbox}=[fill=white, draw=red, shape=rectangle, tikzit category=nodes]
\tikzstyle{orangebox}=[fill=white, draw={rgb,255: red,255; green,128; blue,0}, shape=rectangle]

\tikzstyle{arrow_label}=[->]
\tikzstyle{redarrow}=[draw=red, ->]
\tikzstyle{orangearrow}=[tikzit category=arrows, draw={rgb,255: red,255; green,128; blue,0}, ->]

\usepackage{tkz-graph}
\usetikzlibrary{cd}
\usepackage{hyperref}
\usepackage{amsmath}
\usepackage{xstring}
\acrodef{LR}{Language Resource}
\acrodef{LRT}{Language Resource and Technologies}
\acrodef{LRS}{Language Resource Switchboard}
\acrodef{RI}{Research Infrastructure}
\acrodef{NLP}{Natural Language Processing}
\acrodef{EIF}{European Interoperability Framework}
\acrodef{HIMSS}{Healthcare Information and Management Systems Society, Inc.}
\acrodef{LT}{Language Technologies}
\acrodef{SSH}{Social Sciences and Humanities}
\acrodef{KAF}{Kyoto Annotation Format}
\acrodef{LMF}{Lexical Markup Framework}
\acrodef{GrAF}{Graph Annotation Format}
\acrodef{TEI}{Text Encoding Initiative}

\begin{document}

\title{A Category Theory Approach to Interoperability}
\author{
 Riccardo Del Gratta \\
  Institute for Computational Linguistics ``A. Zampolli''\\
  National Council of Research\\
  Pisa, Italy \\
  \texttt{riccardo.delgratta@ilc.cnr.it} \\
}
\date{}

\maketitle

\begin{abstract}
In this article, we propose a Category Theory approach to (syntactic) interoperability between linguistic tools. The resulting category consists of textual documents, including any linguistic annotations, \acs{NLP} tools that analyze texts and add additional linguistic information,
and format converters. Format converters are necessary to make the tools both able to read and to produce different output formats, which is the key to interoperability. The idea behind this document is the parallelism between the concepts
of composition and associativity in Category Theory with the \acs{NLP} pipelines. We show how pipelines of linguistic tools can be modeled into the conceptual framework of Category Theory and we successfully apply this method to two real-life examples.   
\end{abstract} 

\section{Motivation and plan of the paper}
This article does not pretend to rewrite the approach to syntactic interoperability within the chains of linguistic tools. The concepts behind \acs{NLP} suites or platforms such as those described in Appendices, already consider the idea of an exchange format in which to read and write textual data (and linguistic annotations) an essential one. Independently from the design of such suites, they work in a sort of ``cathedra mea, regulae meae''-perspective: they provide plugins, of course, but suggest to use what they provide. 

In the field of \acl{SSH}, however, there are many legacy tools or tools that are used to visualize data and it gets hard to integrate such tools into \acs{NLP} suites or platforms.

This paper, instead, looks at the exchange format and the format  converters (that play the role of the plugins) from an abstract perspective: it shows that syntactic interoperability can be modeled with a Category Theory approach.

Readers unfamiliar with the world of \acl{LRT} can find a small summary in Appendices \ref{app:lrt}, \ref{app:ri}, and \ref{app:la}.

Appendix \ref{app:lrt} describes \aclp{LR}, \acl{LT}, and introduces \ac{NLP} tools, while Appendix \ref{app:ri} delineates the issues of interoperability as they are addressed by International projects and \aclp{RI}.
\acresetall
\section{Introduction}
\label{sec:intro}
The entire idea of interoperability relies on both composition and associativity. Generally speaking, let's suppose we have a system with $3$ \textit{agents}, call them $a0,a1,a2$, which act on a bunch of data $d0$ to produce $d3$ as a result.

Let's suppose again that the agent $ai$ acts on some $di$ to produce $dj$. We can simply formalize the process $d0 \xrightarrow{}d3$ as $d0 \xrightarrow{a0} d1$ then  $d1^\prime \xrightarrow{a1} d2$ finally $d2^\prime \xrightarrow{a2} d3$. We explicitly put a prime sign $^\prime$ to suggest that the case when the output of an agent  is the exact input for another is infrequent. Indeed,  we can apply $a1$ after $a0$ if and only if the data $d1^\prime$ on which $a1$ acts is compatible, to some extent, with the data $d1$ created by $a0$. In other words, $a1$ and $a0$ \textit{speak the same language} in terms of some characteristics of $di$ such as formats, interchanged data and their meaning. We say that when two agents $ai, aj$ \textit{speak the same language} they are \textit{compatible} with. So, if $a1$ and $a0$ are compatible, we can create a new agent, $a3$, simply \textit{putting together} (i.e. composing) $a1$ with $a0$: $d0 \xrightarrow{a3} d2$. Also, if $a2$ and $a1$ behave as $a1$ and $a0$ do, that's to say they are compatible, we can compose $a2$ with $a1$: the new agent $a4$ acts on $d1$ to produce $d3$, $d1 \xrightarrow{a4} d3$. At this point, we have $5$  \textit{compatible agents} $a0,a1,a2,a3,a4$ that we can \textit{associate} in different ways: either $d0 \xrightarrow{a2 \text{ after } a3} d3$ or $d0 \xrightarrow{a4 \text{ after } a0} d3$.

If we substitute the term agent with function, a bunch of data with elements of a Set and the phrase ``\textit{speak the same language} in terms \dots'' with ``restriction on domain and codomain'' we obtain the theory of functions in Sets. 

Well, Category Theory uses the same model. Instead of elements of a Set or a bunch of data, there are \textit{objects}, instead of functions or agents there are \textit{arrows (morphisms)} between \textit{objects}. Moreover, in Category Theory, composition and associativity are key concepts as they are in interoperability. And if we look at the theory of functions in Sets, agents acting on data and at \textit{objects} and \textit{morphisms} in Category Theory, we see that they are very similar. All these similarities form the idea that is behind the paper.

\acresetall
\section{Background}
\label{sec:back}
Interoperability is a general concept, commonly related to systems (in their broadest sense) able to work together without restrictions. As explained in the dedicated website, \url{http://interoperability-definition.info/en/}, interoperability goes beyond the concept of compatibility between systems, since it is based on agreed structures and open standards. In this way each system is compatible with each other limiting, or even avoiding, the preponderance of one system over the others.

Interoperability is widely used in many disciplines, from healthcare to the medical industry; from services for citizens to emergency management; from computer science to proper software interoperability. For example, the  \ac{EIF}\footnote{\url{https://joinup.ec.europa.eu/collection/nifo-national-interoperability-framework-observatory/3-interoperability-layers}} identifies $5$ levels of interoperability: from technical to legal, while the \ac{HIMSS}\footnote{\url{https://www.himss.org/library/interoperability-standards/what-is-interoperability}} refers to $4$: from foundational to organizational. These two subjects cover very different areas: \ac{EIF} covers public services, \ac{HIMSS} healthcare, but both of them underline \textit{syntactic} and \textit{semantic} interoperability. 

Not surprisingly,  when we come to formalize the concept of interoperability within computer systems, these two terms frequently emerge.  Syntactic interoperability is a prerequisite for semantic interoperability and concerns data formats, communication protocols and everything that can be labeled as \textit{structural}. Formats such as XML, SQL dumps, JSON\ldots are the prototypical examples of agreed data structures and form the structural backbone for syntactic interoperability.

Semantic interoperability focuses on the agreement of the meaning of data exchanged. And this is the place where available standards begin to play a key role. This is especially true for \acp{LR}. Many efforts have been directed toward documenting the \acp{LR} that, although different, could be mapped in some way \cite{ide2010,IdePCS09}. Hence the idea of establishing maps between metadata systems and  the use of data categories and controlled vocabularies, \cite{CieriCCLLPIP10}.
In the realm of \aclp{LR} software integration platforms such as GATE \cite{Cun00a}, UIMA\footnote{\url{https://uima.apache.org/}} \cite{UIMA:FERRUCCI:2004,OASIS:UIMA:2009}, European projects,  and  \acp{RI}, such as CLARIN and DARIAH (see Appendices \ref{app:lrt} and \ref{app:ri}) massively use  the concepts of syntactic and semantic interoperability for the (linguistic) services they offer to users. 

In CLARIN, WebLicht offers linguistic chains based on an agreed structure which is sent from one tool to the next one, while the \ac{LRS} connects individual texts with \acs{NLP} tools. Both of them are based on interoperability. 

Interoperability is also important in Computational Philology. In \cite{Springer2016} the authors describe how to re-engineer \aclp{LR} and \acs{NLP} tools as Web Services to address issues of the digital humanists. Interoperability is used to make connections between lexicons, semantic resources, and fine-grained text management.

Category Theory has been applied to different fields\footnote{In section \ref{sec:cattheo}, we report some references on Applied Category Theory.}, from functional programming languages (ML, Haskell \ldots), to physics, logic, chemistry, semantic web,  software design, and linguistics. Category Theory and linguistics are in close combination. For example, \cite{preller_lambek_2007} and \cite{lambek2008word} use Category Theory and Pregroups to model grammar and interactions among words, while \cite{coecke2013lambek} and \cite{coecke2010mathematical} define and update \textit{DisCoCat}, a model that provides compositional semantics for the study of the meanings of sentences in natural languages.
In the field of semantic web \cite{Cafezeiro:2007} and  \cite{AntunesA18} use  concepts from Category Theory and apply them to ontologies: \textit{limit}, \textit{colimit}, \textit{pushout}, and \textit{pullback}, are used to define optimal \textit{morphisms} 
between ontologies so that they can be enriched and merged. Both works present types of research in the field of semantic interoperability.

\cite{healy2000} uses the same concepts for the design of industrial software. They conclude that a formal approach is necessary to create automated software specification, development, and maintenance. 
\acresetall
\section{Category Theory}
\label{sec:cattheo}
Category Theory is a branch of pure mathematics. In some ways, it can be seen as an abstraction of algebraic structures that include a class of \textit{objects} and a class of \textit{arrows} that connect the objects.

Category Theory has a ``dual approach'': one can learn a good deal  on the \textit{objects} by studying \textit{arrows} and, conversely, many things can be said about \textit{arrows} when they are applied to specific \textit{objects}.

Complete materials on Category Theory and Applied Category Theory can be found here \cite{Awodey:2010:CT:2060081}, \cite{mclane1998}, \cite{Baez34405}, and in \cite{2018arXiv180905923B} and \cite{riehl2017category} along with their references.

\subsection{Definition of a Category}
\label{ssec:catdef}
A Category $\mathcal{C}$ is:
\begin{itemize}
    \item A collection of \textit{objects}, $Ob(\mathcal{C})$
    \item For every pair $X,Y$ $\in$ $Ob(\mathcal{C})$, a collection  (even empty)  of \textit{morphisms} ,\textit{arrows}, between \textit{objects}, $Hom_{\mathcal{C}}(X,Y)$.
    \item Additional \textit{Axioms}:
    \begin{itemize}
        \item There must exist an  \textit{identity arrow} which starts and ends on the same object;
        \item The composition of \textit{arrows} must be associative.
    \end{itemize}
    
\end{itemize}

\begin{description}

\item[Composition:] 
$f$ is a morphism from $A$ to $B$,  $g$ from $B$ to $C$:
\[A \xrightarrow{f} B, B \xrightarrow{g} C\] 
Given $f, g$ there must exist $h$ which is the composite of $f$ and $g$:
\begin{equation}
    \label{eq:comp}
    h=f\circ g
\end{equation}
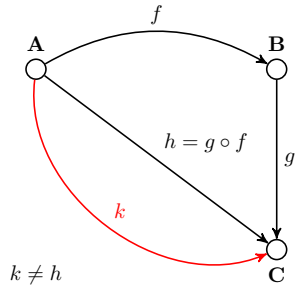
\begin{figure}[H]
    \centering
    \begin{tikzpicture}[>=stealth',semithick,auto,scale=0.8, every node/.style={scale=0.8}]
   
    \tikzstyle{every label}=[font=\bfseries]

    \node[style={empty_node}, label={above:A}] (xa) at (1,1) {};
    \node[style={empty_node}, label={above:B}] (xb) at (5,1) {};
    \node[style={empty_node}, label={below:C}] (xc) at (5,-2) {};
    \node[label={below:$k \neq h$}] (xd) at (1,-2) {};
    \path[->]   (xa)    edge[bend left=30]                node{$f$}       (xb)
                (xb)        edge                node        {$g$}  (xc)
                (xa)    edge  node        {$h=g\circ f$}  (xc) 
                (xa)    edge[bend right=60, red]  node        {$k$}  (xc);

\end{tikzpicture}
    \caption{Composition in Category Theory.}\label{fig:comp}
\end{figure}

The morphism $k$ in Figure \ref{fig:comp} can be any morhism between $A$ and $C$ and not necessarily the composite $f$ and $g$.
\item[Identity: ] The \textit{identity} morphism  is defined as $X \xrightarrow{id_X} X$; when we apply the composition to $f$ as in Figure \ref{fig:id} 

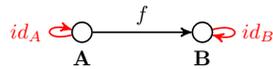
\begin{figure}[H]
    \centering
    \begin{tikzpicture}[>=stealth',semithick,auto,scale=0.8, every node/.style={scale=0.8}]
   
    \tikzstyle{every label}=[font=\bfseries]

    \node[style={empty_node}, label={below:A}] (xa) at (1,1) {};
    \node[style={empty_node}, label={below:B}] (xb) at (3,1) {};
    \path[->]   (xa)    edge node{$f$}       (xb)
                (xa)    edge[loop left=50, red]  node        {$id_{A}$}  (xa)
                (xb)    edge[loop right=50, red]  node        {$id_{B}$}  (xb);

\end{tikzpicture}
    \caption{Identity and Composition.}\label{fig:id}
\end{figure}
we obtain
\begin{equation}
    \label{eq:ind}
    f \circ id_A = f = id_B \circ f
\end{equation}
\item[Associativity: ] Composition leads to associativity, see Figure \ref{fig:assoc}. In the sense that for $\forall f,g,h$ in $Hom_{\mathcal{C}}(X,Y)$ there must be:
\begin{equation}
    \label{eq:assoc}
    h\circ (g\circ f)= (h\circ g) \circ f = h \circ g \circ f
\end{equation}
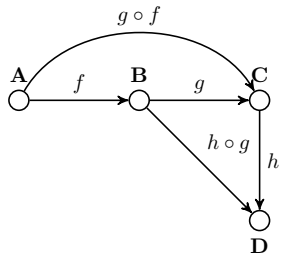
\begin{figure}[H]
    \centering
\begin{tikzpicture}[>=stealth',semithick,auto,scale=0.8, every node/.style={scale=0.8}]
   
    \tikzstyle{every label}=[font=\bfseries]

    \node[style={empty_node}, label={above:A}] (xa) at (1,1) {};
    \node[style={empty_node}, label={above:B}] (xb) at (3,1) {};
    \node[style={empty_node}, label={above:C}] (xc) at (5,1) {};
    \node[style={empty_node}, label={below:D}] (xd) at (5,-1) {};
    \path[->]   (xa)    edge                node{$f$}       (xb)
                (xb)    edge                node        {$g$}  (xc)
                (xc)    edge                node        {$h$}  (xd)
                (xa)    edge[bend left=60]  node        {$g\circ f$}  (xc) 
                (xb)    edge  node        {$h \circ g$}  (xd);

\end{tikzpicture}
    \caption{Composition leads to Associativity.}\label{fig:assoc}
\end{figure}

\end{description}
\acresetall
\section{Interoperability and Category Theory}
\label{sec:intict}
Composition and associativity are important concepts in Category Theory. Just as they are in interoperability.

\begin{figure}[H]
    \centering
\begin{tikzpicture}[scale=1.0, every node/.style={scale=0.8}]
	\begin{pgfonlayer}{nodelayer}
		\node [style=blackbox] (0) at (-1, 0) {$t_{1}$};
		\node [style=blackbox] (1) at (1, 0) {$t_{2}$};
		\node [style=redbox] (2) at (0, -1) {$t_{1} \circ t_{2}$};
		\node [style=none] (3) at (-2, 0) {};
		\node [style=none] (4) at (2, 0) {};
	\end{pgfonlayer}
	\begin{pgfonlayer}{edgelayer}
		\draw [style=redarrow, bend right, looseness=1.25] (2) to (4);
		\draw [style=redarrow, bend right] (3) to (2);
		\draw [style={arrow_label}] (0) to (1);
		\draw [style={arrow_label}] (1) to (4.center);
		\draw [style={arrow_label}] (3.center) to (0);
	\end{pgfonlayer}
\end{tikzpicture}

    \caption{Composition of linguistic tools.}\label{fig:int1}
\end{figure}
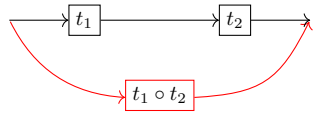

The comparison between the composition in Category Theory and interoperability  of linguistic tools is quite immediate. When we require two tools to be interoperable we mean exactly that the output of the first tool ($t_1$) is the input of the second one ($t_2$). And that these two tools can be grouped to get a more complex, \textit{composite}, tool ($t_2\circ t_1$) that provides the same results, as reported in Figure \ref{fig:int1}.

Similarly, in the case of more tools, the processing pipeline(s) can proceed in different ways: we can obtain the same results using either atomic or \textit{composite} tools\footnote{By atomic tools we mean tools that go directly from $A$ to $B$: $A \xrightarrow{t} B$; by composite, tools that need an intermediate $C$ to go from $A$ to $B$: $A \xrightarrow{t_{C}} B$}  as in Figures \ref{fig:int2} and \ref{fig:int21}.
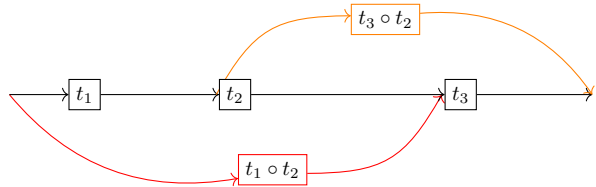
\begin{figure}[htbp]
    \centering
\begin{tikzpicture}[scale=1.0, every node/.style={scale=0.8}]
	\begin{pgfonlayer}{nodelayer}
		\node [style=blackbox] (0) at (-1, 0) {$t_{1}$};
		\node [style=blackbox] (1) at (1, 0) {$t_{2}$};
		\node [style=redbox] (2) at (1.5, -1) {$t_{1} \circ t_{2}$};
		\node [style=none] (3) at (-2, 0) {};
		\node [style=blackbox] (5) at (4, 0) {$t_{3}$};
		\node [style=none] (6) at (3.75, 0) {};
		\node [style=none] (7) at (0.75, 0) {};
		\node [style=none] (8) at (5.75, 0) {};
		\node [style=orangebox] (10) at (3, 1) {$t_{3} \circ t_{2}$};
	\end{pgfonlayer}
	\begin{pgfonlayer}{edgelayer}
		\draw [style=redarrow, bend right] (3.center) to (2);
		\draw [style={arrow_label}] (0) to (1);
		\draw [style={arrow_label}] (3.center) to (0);
		\draw [style=redarrow, bend right, looseness=1.25] (2) to (6.center);
		\draw [style={arrow_label}] (5) to (8.center);
		\draw [style={arrow_label}] (1) to (5);
		\draw [style=orangearrow, bend left=30] (10) to (8.center);
		\draw [style=orangearrow, bend left=30, looseness=1.25] (7.center) to (10);
	\end{pgfonlayer}
\end{tikzpicture}

    \caption{Associativity.}\label{fig:int2}
\end{figure}
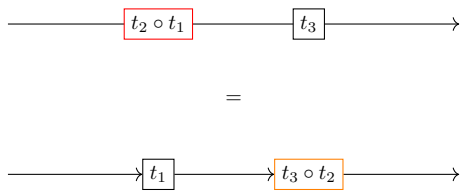
\begin{figure}[H]
    \centering
\begin{tikzpicture}[scale=1.0, every node/.style={scale=0.8}]
	\begin{pgfonlayer}{nodelayer}
		\node [style=none] (0) at (0, 0) {$=$};
		\node [style=none] (1) at (-3, 1) {};
		\node [style=none] (2) at (3, 1) {};
		\node [style=none] (3) at (-3, -1) {};
		\node [style=none] (4) at (3, -1) {};
		\node [style=redbox] (5) at (-1, 1) {};
		\node [style=blackbox] (6) at (1, 1) {};
		\node [style=blackbox] (7) at (-1, -1) {$t_1$};
		\node [style=orangebox] (8) at (1, -1) {$t_{3} \circ t_{2}$};
		\node [style=redbox] (9) at (-1, 1) {$t_{2} \circ t_{1}$};
		\node [style=blackbox] (10) at (1, 1) {$t_{3}$};
	\end{pgfonlayer}
	\begin{pgfonlayer}{edgelayer}
		\draw [style={arrow_label}] (6) to (2.center);
		\draw [style={arrow_label}] (3.center) to (7);
		\draw [style={arrow_label}] (7) to (8);
		\draw [style={arrow_label}] (8) to (4.center);
		\draw [style={arrow_label}] (1.center) to (5);
		\draw [style={arrow_label}] (5) to (6);
	\end{pgfonlayer}
\end{tikzpicture}

    \caption{Associativity as an equation.}\label{fig:int21}
\end{figure}
Figures \ref{fig:int1} and \ref{fig:int2} are the diagrammatic counterpart of Equations \ref{eq:comp} and \ref{eq:assoc} respectively.\\

\acresetall
\section{Building the Category}
\label{sec:build}
It is therefore natural to identify the objects of the category with the textual documents\footnote{Henceforth we use document instead of textual document.} to be analyzed and the morphisms with the linguistic applications between them.

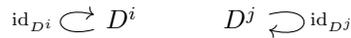
\begin{figure}[H]
    \centering
   \begin{tikzcd}
D^i \arrow[loop left]{r}{\mathrm{id}_{D^{i}}} & 
D^j \arrow[loop right]{r}{\mathrm{id}_{D^{j}}} 
\end{tikzcd}
    \caption{Identities.} \label{fig:tii}    
    
\end{figure}

According to Section \ref{sec:cattheo} we can define a category $\mathcal{C}$ as follows:
\begin{itemize}
    \item The collection of \textit{objects}, $Ob(\mathcal{C})$, consists of all  documents ($D^{i}$,$D^{ii}$,$D^{iii}$,$D^{iv}$ \ldots) that can be processed by a linguistic application;
    \item  The $Hom_{\mathcal{C}}(D^{i},D^{j})$, the collection of morphisms, is the set of any linguistic application which consumes $D^{i}$ and produces $D^{j}$ as in Figure \ref{fig:tij}.
\end{itemize}
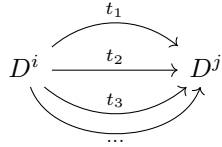
\begin{figure}[H]
    \centering
\begin{tikzcd}
D^i \arrow[rr, "t_1", bend left=40] \arrow[rr, "t_2"] \arrow[rr, "t_3", bend right=40] \arrow[rr, "..."', bend right=70] &  & D^j
\end{tikzcd}
    \caption{Tools from $D^i$ to $D^j$.} \label{fig:tij}    
    
\end{figure}
 The \textit{identity} morphism is a dummy tool that consumes and returns the same  document. These morphisms can be called  the ``do-nothing'' tools in analogy with the identity function in Haskell, $ id :: x \rightarrow x$, that returns its argument unchanged, or with the \textit{pass} statement in Python. There must be an identity morphism for every  document $D^i$, see Figure \ref{fig:tii}.
Composition and associativity are reported in Figure \ref{fig:tijkm}: 

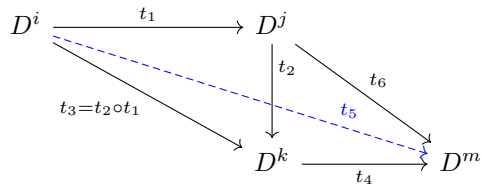
\begin{figure}[H]
    \centering
\begin{tikzcd}
D^i \arrow[rrr, "t_1"] \arrow[rrrdd, "t_3 = t_2 \circ t_1"'] \arrow[rrrrrdd, "t_5", dashed, near end,blue] &  &  & D^j \arrow[dd, "t_2", near start] \arrow[rrdd, "t_6"] &  &     \\
                                                                                                      &  &  &                                                       &  &     \\
                                                                                                      &  &  & D^k \arrow[rr, "t_4"']                                &  & D^m
\end{tikzcd}

    \caption{Composition and associativity.} \label{fig:tijkm}    
    
\end{figure}
A  document $D^i$ is processed to obtain $D^m$ in different ways: directly using  $t_5$ (the {\color{blue}blue dashed} line); composing $t_4$ with $t_3$ (which is actually $t_2 \circ t_1$); composing $t_6$ and $t_1$ and doing the same with $t_4$, $t_2$, and $t_1$.
It seems we are done. Anyway, we are not.

\acresetall
\section{More thoughts on texts and \acs{NLP} tools}
\label{sec:more}
What are the \acs{NLP} tools? Simply put, an \acs{NLP} tool is a software able to process natural language data and perform linguistic operations on them. Here, by language data we mean any collection of  documents. These can be simple documents, written in natural languages, collections of words, annotated documents (e.g. documents which already contain linguistic information), formatted documents (for example in tabbed fields) and so on. Both definitions are not exhaustive (see Appendices \ref{app:lrt}, \ref{app:ri}, and \ref{app:la} for a brief introduction) but, for the scope of the article, we don't need to formally and exhaustively define \acs{NLP} tools and  documents: it suffices to say that \acs{NLP} tools perform linguistic operations on documents.\footnote{We provide a generic definition of documents and tools in Section \ref{sec:ttcl}.}\\
As a consequence, \acs{NLP} tools are classified according to the linguistic operation(s) they perform on  documents: there are part-of-speech taggers, which assign morphological features such as \texttt{VERB}, \texttt{NOUN}\ldots to words, lemmatizers which assign to inflected forms their dictionary entry (e.g. from \textit{loves} to \textit{love}), language identifiers, parsers, word sense disambiguators which pick up the right sense of a word (e.g in the sentence ``I went to the bank yesterday to get some money'', bank is the financial institution and not the sloping land of a river) and so on.

The most important aspect is that not all  documents can be processed by any \acs{NLP} tool. For example, there could be a lemmatizer ($t_{l_1}$) which reads a list of words ($D^1$) and assign the lemma to each of them (without considering the structure of the text). But there could be a different lemmatizer ($t_{l_2}$) which reads plain text sentence by sentence ($D^2$) because its algorithm reads words in context. And there could be a third lemmatizer ($t_{l_3}$) which needs the part of speech of the words ($D^3$) to assign lemmas. 

\begin{figure}[H]
    \centering
    \begin{tikzpicture}[>=stealth',semithick,auto,scale=1.0, every node/.style={scale=1.0}]
   
    \tikzstyle{every label}=[font=\bfseries]

    \node[style={empty_node}, label={above:$D^{4}$}] (xa) at (0,2) {};
    \node[style={empty_node}, label={below:$D^{1}$}] (xb) at (-2,0) {};
    \node[style={empty_node}, label={below:$D^{2}$}] (xc) at (0,0) {};
    \node[style={empty_node}, label={below:$D^{3}$}] (xd) at (2,0) {};
    \path[->]   (xb)    edge[]                node {\color{red}{$t_{l_{1}}$}}       (xa)
                (xc)    edge[]                node  {\color{red}{$t_{l_{2}}$}}       (xa)
                (xd)    edge[]                node  {\color{red}{$t_{l_{3}}$}}       (xa);

\end{tikzpicture}
 \caption{Multiple tools producing the same result.} \label{fig:t4}    
    
\end{figure}
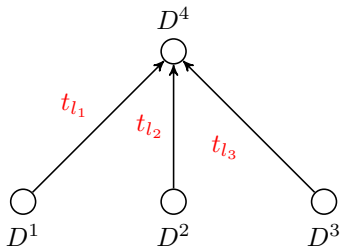

\begin{figure}[H]
    \centering
    \begin{tikzpicture}[>=stealth',semithick,auto,scale=1.0, every node/.style={scale=1.0}]
   
    \tikzstyle{every label}=[font=\bfseries]

    \node[style={empty_node}, label={above:$D^{4}$}] (xa) at (0,2) {};
    \node[style={empty_node}, label={below:$D^{1}$}] (xb) at (-2,0) {};
    \node[style={empty_node}, label={below:$D^{2}$}] (xc) at (0,0) {};
    \node[style={empty_node}, label={below:$D^{3}$}] (xd) at (2,0) {};
    \path[->]   (xb)    edge[]                node {\color{red}{$t_{l_{1}}$}}       (xa)
                (xc)    edge[]                node  {\color{red}{$t_{l_{2}}$}}       (xa)
                (xd)    edge[]                node  {\color{red}{$t_{l_{3}}$}}       (xa);
                
     \path[dotted,->]  (xc)    edge[]                node {\color{red}{$t_{p_{1}}$}}       (xd);
\end{tikzpicture}
 \caption{$D^2$  as source for $D^3$.} \label{fig:t5}    
    
\end{figure}
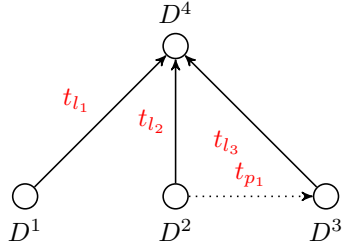

The fact that a tool needs specific input is not surprising and it is indeed well known, especially in \acp{RI} as  \cite{Odijk2018DiscoveringSR} reports.
In the example above, the tools $t_{l_i}$ produce the same result (text with lemmas) starting from three different inputs. If we call $D^4$ the \textit{text\_with\_lemmas}, the situation goes as in Figure \ref{fig:t4}.
The same example tells us that there are no substantial differences between the documents  $D^1$ and $D^2$, apart from their format (list of words vs. plain text), while $D^3$ contains additional information (the part of speech). It is natural to suppose that $D^3$ can be obtained from (for instance) $D^2$ applying a part-of-speech tagger, see Figure \ref{fig:t5}.
\acresetall
\section{Tuning the Category}
\label{sec:ttcl}
According to \ref{sec:more}, the category $\mathcal{C}$ sketched in Section \ref{sec:build} is not complete. We have to take into consideration that i)  documents with the same linguistic information can have different formats and ii) the final result can be obtained from documents containing different linguistic information.

\subsection{Category Objects}
\label{ssec:catobj}
In Section \ref{sec:build}, we defined the objects, $Ob(\mathcal{C})$, as the  documents ($D^{i}$,$D^{ii}$,$D^{iii}$,$D^{iv}$\ldots) that can be processed by linguistic applications. From Section \ref{sec:more} we learned that such  documents are more complex than the ones covered by the definition. 
We define the documents $D^i$ as follows:
\begin{equation}
    \label{eq:thetext}
    D \equiv D(c,f,\{a_1\ldots a_n\})
\end{equation}
where $c$ is the content (e.g. the text), $f$ the format and $\{a_1, \dots, a_n\}$ the set of additional linguistic annotations (if any). From  \ref{eq:thetext}, follows the definition for the \textit{initial}  document:
\begin{equation}
    \label{eq:thetext0}
    D^0 \equiv D(c,f_{p},\{\emptyset\})
\end{equation}
where $f_{p}$ is the format corresponding to plain text  and $\{\emptyset\}$ means that there is no additional linguistic annotation. In definitions \ref{eq:thetext} and \ref{eq:thetext0}, $f_{(p)}$ is how both  $c$ and $\{a_1, \dots, a_n\}$ are serialized in a data structure.

\subsection{Category Morphisms}
\label{ssec:catmor}
When a document $D^i$ is analyzed with  an \acs{NLP} tool $\tilde t_{ij}$, a  document $D^j$ is  produced:
\begin{equation}
    \label{eq:genmor}
    \tilde t_{ij} \coloneqq D^i(c_i,f_i,\{a_i\})\xrightarrow{} D^j(c_j,f_j,\{a_j\})
\end{equation}
Equation \ref{eq:genmor} represents the most general morphism that acts on a document $D_i$, in the sense that $\tilde t_{ij}$ modifies all the three components of $D_i$, the content $c$, the format $f$, and the annotation set $\{a\}$. The collection of such $\tilde t$ is the hom-set of the Category:
\[\tilde t_{ij} \in Hom_{\mathcal{C}}(D^{i},D^{j})\]

In Computational Linguistics, depending on the specific tool $\tilde t$, we expect that $D^i$ and $D^j$ may (or may not) differ for the content, the format, and the annotation set. For example, if $\tilde t$ is a named entity extractor and  $D^i$ is the initial text $D^0$, $D^j$ may either have the same content as $D^0$ with an additional layer of stand-off annotations consisting of words and named entities or be a simple list of extracted named entities showing no trace of the original content. In addition, the output of $\tilde t$, $(D^j)$, can be serialized in XML which might not be the original format of $D^0$. The \acs{NLP} tool $\tilde t$ is ``something'' acting on the format, ``something'' on the  content and ``something'' on the annotation set. 

Consequently, we can proceed by defining such ``something''. As a useful simplification, one can look at the document in \ref{eq:thetext} as the Cartesian product\footnote{Such documents are called \textit{separable}} of the content $c$, the format $f$, and the annotation set $\{a\}$:
\begin{equation}
    \label{eq:thetext1}
    D^i(c_i,f_i,\{a_i\}) = c_i \times f_i\times \{a_i\}
\end{equation}
Unfortunately, this is quite never the case since the content, format, and annotation sets are closely interconnected\footnote{For example, in KAF (see Section \ref{sec:rle}), the content in embedded in the format.} requiring the more general definition of morphisms as in \ref{eq:genmor}, but can help to understand the sub-classes of morphisms defined below. \\
We can adopt the following definition of format converters: a format \textit{converter} is an application that connects two documents and leaves $c$ and $\{a_1, \dots, a_n\}$ unchanged while moving from format $f_i$ to format $f_j$.
\begin{equation}
    \label{eq:theconv}
    c_{ij} \coloneqq D^i(c,f_i,\{a_1\ldots a_n\}) \xrightarrow{} D^j(c,f_j,\{a_1\ldots a_n\})
\end{equation}
Definition \ref{eq:theconv}, when applied to the documents in \ref{eq:thetext1}, means that the format converters act as \textit{identities} on the content $c$ and the annotation set $\{a\}$:
\begin{equation}
    \label{eq:theconv1}
    c_{ij} \coloneqq Id_c \times f_{ij} \times Id_{\{a\}}
\end{equation} 
where $f_{ij}$ is an application that changes the format form $f_i$ to $f_j$.

Following Appendix \ref{app:lrt}, an \acs{NLP} tool  is an application that connects two documents and may change both content $c$ and annotations $\{a_1, \dots, a_n\}$ leaving the format $f$ unchanged.
\begin{equation}
    \label{eq:thetool}
    t_{ij} \coloneqq D^i(c_i,f,\{a_1\ldots a_k\}) \xrightarrow{} D^j(c_j,f,\{a_1\ldots a_n\})
\end{equation}

This is a strong position and is openly in contrast with the fact that \acs{NLP} tools consume specific inputs and produce specific outputs, as reported in Section \ref{sec:more}; definition \ref{eq:thetool}, when applied to the documents in \ref{eq:thetext1}, means that the \acs{NLP} tools act as \textit{identities} on the format $f$:
\begin{equation}
    \label{eq:thetool1}
    t_{ij} \coloneqq g_{ij} \times Id_f \times a_{ij}
\end{equation}
where $h_{ij}, a_{ij}$ are applications that change the either the content or the annotation sets or both.  

Neither definitions \ref{eq:thetool} (or \ref{eq:thetool1})  nor definitions \ref{eq:theconv} (or \ref{eq:theconv1}) capture the actual issues related to the implementation of \acs{NLP} tools. It is often the case that a tool does ``something'' on the format too, that is to say behave as $\tilde t$ does in definition \ref{eq:genmor}. Many times the tools are designed to be a sort of \textit{composition} of a converter and a \acs{NLP} tool: indeed, when one tool is concretely implemented, it is distributed as a whole, even if it is logically divided in format converter and proper tool, see references in Appendix \ref{app:lrt}. 

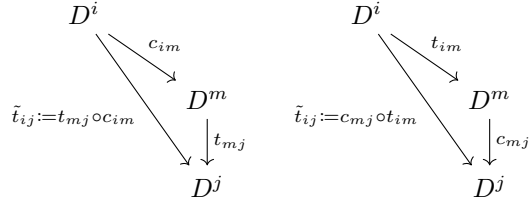
\begin{figure}[htbp]
    \centering
    \begin{subfigure}[h]{0.3\textwidth}
             \begin{tikzcd}
D^i \arrow[rd, "c_{im}"] \arrow[rdd, "\tilde t_{ij} \coloneqq  t_{mj} \circ c_{im}"'] &                         \\
                                                             & D^m \arrow[d, "t_{mj}"] \\
                                                             & D^j                    
\end{tikzcd}
        \caption{the tool $t_{mj}$ is executed after the converter $c_{im}$.} \label{fig:tafterc}
    \end{subfigure}
    \begin{subfigure}[h]{0.3\textwidth}
                    \begin{tikzcd}
D^i \arrow[rd, "t_{im}"] \arrow[rdd, "\tilde t_{ij} \coloneqq  c_{mj} \circ t_{im}"'] &                         \\
                                                             & D^m \arrow[d, "c_{mj}"] \\
                                                             & D^j                    
\end{tikzcd}
       \caption{the converter $c_{mj}$ is executed after the tool $t_{im}$.} \label{fig:caftert}
    \end{subfigure}

    \caption{Combination between tools and converters.}\label{fig:taftercaftert}
\end{figure}

In this article, with the specific aim of interoperability, we look at the $\tilde t$ tools (the morphisms defined in \ref{eq:genmor}) as a sort of either the \textit{composition} of a tool $t_{ij}$ and a converter $c_{ij}$ or vice-versa, see Figures \ref{fig:tafterc} and \ref{fig:caftert}.

where  $\tilde t$ is defined as either $c \circ t$ or $t \circ c$. The document $D^m$ in
Figure \ref{fig:taftercaftert} is a document which has the same content but different format of $D^i$ (Figure \ref{fig:tafterc}) and used as input document for $t_{mj}$. Conversely, $D^m$ has different content (for instance) (Figure but same format  of $D_i$ \ref{fig:caftert}) and is eventually transformed to $D^j$.

According to the definition of $Ob(\mathcal{C})$, both $D^i$ and $D^j$ in equations \ref{eq:theconv} and \ref{eq:thetool} belong to $Ob(\mathcal{C})$ with the additional constraint that $D^i$ $\neq$ $D^j$ and the $Hom_{\mathcal{C}}(D^{i},D^{j})$ contains (pure) format converters, tools that act on content and/or annotation set, and tools that act on both format and content and/or annotation set (the $\tilde t$ in \ref{eq:genmor}).

In a categorical perspective, we have to verify that composition between tools and converters as defined in \ref{eq:theconv} and \ref{eq:thetool} make sense. \\
A converter $c_{ij}$ composed with a converter $c_{jk}$ is a converter $c_{ik}$. From 
\[c_{ij}\coloneqq D_i \xrightarrow[]{} D_j \text{ and } c_{jk}\coloneqq D_j \xrightarrow[]{} D_k \text{ follows }\]
\[c_{ik}\coloneqq D_i \xrightarrow[]{} D_k\]
Analogously, a tool $t_{ij}$ composed with a tool $t_{jk}$ is a tool $t_{ik}$. From 
\[t_{ij}\coloneqq D_i \xrightarrow[]{} D_j \text{ and } t_{jk}\coloneqq D_j \xrightarrow[]{} D_k \text{ follows }\]
\[t_{ik}\coloneqq D_i \xrightarrow[]{} D_k\]
While the composition of  $c_{ij}$ and $t_{ij}$ and vice-versa are the $\tilde t$ in \ref{eq:genmor}. Thus, $c_{ij}, t_{ij}, \tilde t_{ij} \in Hom_{\mathcal{C}}(D^{i},D^{j})$.

\subsection{Category Axioms}
\label{ssec:axioms}
Because of these new definitions, the three axioms, identity, composition and associativity have to be revised.
\begin{description}
\item[Identity:] identity is still the ``do-nothing'' tool, but such tool does nothing on content, format and annotation set:
\begin{tikzcd}
D^i \arrow[loop left]{r}{\mathrm{id}_{D^{i}}} & 
\end{tikzcd}\\\ 
\[id_{D^i}(c)=c, id_{D^i}(f)=f,id_{D^i}(\{a_k\})=\{a_k\} \]
\item[Composition:] in Section \ref{sec:more}, we explained that a linguistic result can either be obtained from documents with different formats or from documents with different annotation sets, and this is closely correlated to the tool and its input/output restrictions. However, thanks to our definitions of converters (\ref{eq:theconv}) and tools (\ref{eq:thetool}), we know how to address this issue. Figure \ref{fig:compmore} shows a case when a tool $t_{im}$ consumes $D^i$ to produce $D^m$, but $D^m$ can not be provided to $t_{mk}$ to produce $D^k$.
\begin{figure}[H]
\centering
\begin{tikzcd}
D^i \arrow[rd, "t_{im}"] &                            &     \\
                         & D^m \arrow[d, "?", dashed] &     \\
                         & D^j \arrow[r, "t_{jk}"]    & D^k
\end{tikzcd}
\caption{The tool $t_{mk}$ can not act on $D^m$.}\label{fig:compmore}

\end{figure}
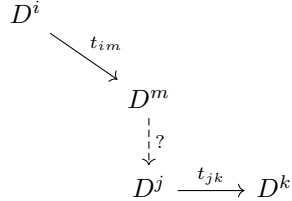
If $D^m$ and $D^j$ differ for their formats, we can apply a converter $c_{mj}$ to obtain $D^j$ and then provide  $D^j$ to $t_{jm}$ to obtain  $D^k$, see Figure \ref{fig:needc}. Otherwise, If $D^m$ and $D^j$ differ for their annotation sets, we can apply a tool $t_{mj}$ to obtain $D^j$ and then provide  $D^j$ to $t_{jm}$ to obtain  $D^k$, see Figure \ref{fig:needt}\footnote{More realistic cases when the various $D^i$ differ for format \textit{and} annotation sets are managed similarly, but are pictorially more complex.}.
\begin{figure}[htbp]
    \centering
    \begin{subfigure}[h]{0.3\textwidth}
             \begin{tikzcd}
D^i \arrow[rd, "t_{im}"] &                            &     \\
                         & D^m \arrow[d, "c_{mj}"] &     \\
                         & D^j \arrow[r, "t_{jk}"]    & D^k
\end{tikzcd}
        \caption{$D^m$ and $D^j$ differ for their formats.} \label{fig:needc}
    \end{subfigure}
    \begin{subfigure}[h]{0.3\textwidth}
                    \begin{tikzcd}
D^i \arrow[rd, "t_{im}"] &                            &     \\
                         & D^m \arrow[d, "t_{mj}"] &     \\
                         & D^j \arrow[r, "t_{jk}"]    & D^k
\end{tikzcd}
      \caption{$D^m$ and $D^j$ differ for their annotation sets.} \label{fig:needt}
    \end{subfigure}

    \caption{Composition involving converters and tools.}\label{fig:needct}
\end{figure}
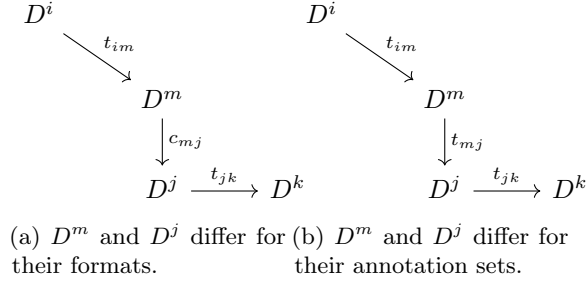

In both cases, if $D^{m\prime}$ is $D^{m}$ transformed, diagrams in Figure \ref{fig:needct} are rendered as in Figure \ref{fig:comptilde}.
\begin{figure}[H]
\centering
\begin{tikzcd}
D^i \arrow[r, "\tilde t_{im\prime}"] \arrow[rr, "\tilde t_{ik}:= \tilde t_{m\prime k} \circ \tilde t_{im\prime} "', bend right] & D^{m\prime} \arrow[r, "\tilde t_{m\prime k} "] & D^k
\end{tikzcd}
\caption{Tools $\tilde t$ ensure composition.}\label{fig:comptilde}
\end{figure}
\item[Associativity:] it follows from Figure \ref{fig:assoctilde}:
\begin{figure}[H]
\centering
\begin{tikzcd}
D^i \arrow[r, "\tilde t_{im}"] \arrow[rr, "\tilde t_{ik}:= \tilde t_{mk} \circ \tilde t_{im} "', bend right] & D^{m} \arrow[r, "\tilde t_{mk} "] \arrow[rr, "\tilde t_{mk}:= \tilde t_{kl} \circ \tilde t_{mk}", bend left=49] & D^k \arrow[r, "\tilde t_{kl}"] & D^l
\end{tikzcd}
\caption{Tools $\tilde t$ ensure composition.}\label{fig:assoctilde}
\end{figure}
From Figure \ref{fig:assoctilde} we have the usual association rule:
\[(\tilde t_{kl} \circ \tilde t_{mk})\circ \tilde t_{im} = \tilde t_{kl}\circ (\tilde t_{mk} \circ \tilde t_{im}) \]
which makes sense thanks to the fact that  $\tilde t \in Hom_{\mathcal{C}}(D^{i},D^{j})$ by construction.
\end{description}
\acresetall
\section{Real-Life Examples}
\label{sec:rle}
The authors in \cite{delgrattacac2019} described the integration of a set of \acs{NLP} tools into  WebLicht and \acl{LRS} and reviewed the encountered interoperability issues. On one hand, WebLicht is a chain of tools, and this implies that \acs{NLP} tools must accept constraints on their input/output formats to be integrated into WebLicht: namely, they have to consume/produce valid TCF documents. On the other hand, \ac{LRS} connects  documents with \acs{NLP} tools via their input format\footnote{Truth be told, \ac{LRS} suggests tools according to the \textit{mime-type}, which is a bit stronger than the format only, of the incoming  documents.}. When we come to manage the integration of \acs{NLP} tools into chains such as WebLicht and infrastructural services as \ac{LRS}, syntactic interoperability emerges. But, at least at the beginning, it can be restricted to conversion issues that are managed with the help of \textit{ad-hoc} wrappers able to connect one  document $D^i$ to another $D^j$. We call such wrappers $W_{ij}$.

\begin{figure}[H]
    \centering
\begin{tikzpicture}[scale=0.8, every node/.style={scale=0.8}]
	\begin{pgfonlayer}{nodelayer}
		\node [style=blackbox] (0) at (0, 0) {$ W_{ij}$};
		\node [style=none] (1) at (-1, 0) {$D^i$};
		\node [style=none] (2) at (1, 0) {$D^j$};
	\end{pgfonlayer}
	\begin{pgfonlayer}{edgelayer}
		\draw [style={arrow_label}] (1) to (0);
		\draw [style={arrow_label}] (1) to (2);
	\end{pgfonlayer}
\end{tikzpicture}
    \caption{Process of wrapping.}\label{fig:wrp1}
\end{figure}
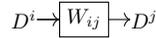

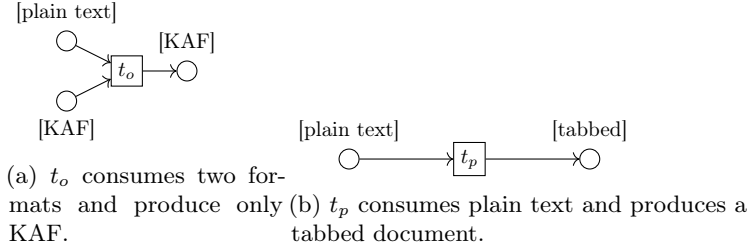
\begin{figure}[htbp]
    \centering
    \begin{subfigure}[b]{0.3\textwidth}
            \begin{tikzpicture}[scale=0.8, every node/.style={scale=0.8}]
	\begin{pgfonlayer}{nodelayer}
		\node [style=blackbox] (0) at (0, 0) {$t_{o}$};
		\node [style={empty_node}, label={above:[plain text]}] (1) at (-1, 0.5) {};
		\node [style={empty_node},label={below:[KAF]}] (2) at (-1, -0.5) {};
		\node [style={empty_node},label={above:[KAF]}] (3) at (1, 0) {};
	\end{pgfonlayer}
	\begin{pgfonlayer}{edgelayer}
		\draw [style={arrow_label}] (1) to (0);
		\draw [style={arrow_label}] (2) to (0);
		\draw [style={arrow_label}] (0) to (3);
	\end{pgfonlayer}
\end{tikzpicture}
        \caption{$t_o$ consumes two formats and produce only KAF.}\label{fig:to}
    \end{subfigure}
    \begin{subfigure}[b]{0.5\textwidth}
 
\begin{tikzpicture}[scale=0.8, every node/.style={scale=0.8}]
	\begin{pgfonlayer}{nodelayer}
		\node [style=blackbox] (0) at (0, 0) {$t_{p}$};
		\node [style={empty_node}, label={above:[plain text]}] (1) at (-2, 0) {};
		\node [style={empty_node},label={above:[tabbed]}] (2) at (2, 0) {};
	\end{pgfonlayer}
	\begin{pgfonlayer}{edgelayer}
		\draw [style={arrow_label}] (1) to (0);
		\draw [style={arrow_label}] (0) to (2);
	\end{pgfonlayer}
\end{tikzpicture}
    \caption{$t_p$ consumes plain text and produces a tabbed document.}\label{fig:tp}
    \end{subfigure}
   
 
    \caption{Native tools with their input/output restrictions.}\label{fig:native}
\end{figure}

$W_{ij}$ can be simplified as a box which receives  documents in inputs and produces new (annotated)  document in output, see Figure \ref{fig:wrp1}.
In \cite{delgrattacac2019}, the wrapper $W_{ij}$ is built around two native tools, $t_o,t_p$, which consume and produce specified formats, see Figure \ref{fig:native}.

More precisely, $t_o$ consumes either plain or KAF \cite{KAF} formats producing KAF (Figure \ref{fig:to}); $t_p$ reads plain texts and produces a tabbed output (Figure \ref{fig:tp}).
The final objective of $W_{ij}$ is to make the native tools able to accept either plain or TCF or KAF  documents as input and provide either tabbed or TCF or KAK  documents as output. Both tools $t_o$ and $t_p$ are tokenizers, therefore the produced output contains the tokenization of the input documents as its annotation set. In addition, $t_p$ does not keep track of the original content, while $t_o$ does.
\subsection{Mapping the process onto Category Theory}
\label{ssec:mapct}
\begin{figure}[htbp]
    \centering
    \begin{subfigure}[b]{0.3\textwidth}
            \begin{tikzpicture}[scale=0.8, every node/.style={scale=0.8}]
	\begin{pgfonlayer}{nodelayer}
		\node [style=blackbox] (0) at (0, 0) {$t_{o}$};
		\node [style={empty_node}, label={above:[plain text]}] (1) at (-1, 0.5) {};
		\node [style={empty_node},label={below:[KAF]}] (2) at (-1, -0.5) {};
		\node [style={empty_node},label={above:[KAF]}] (11) at (1, 0) {};
		\node [style=none] (3) at (-2.5, -1.5) {};
		\node [style=none] (4) at (-2.5, 1.5) {};
		\node [style=none] (5) at (2.5, -1.5) {};
		\node [style=none] (6) at (2.5, 1.5) {};
		\node [style=none] (7) at (2.5, 0) {};
		\node [style=none] (8) at (-2.5, 0) {};
		\node [style=none] (9) at (-3.5, 0) {$D^i_{f_i}$};
		\node [style=none] (10) at (3.5, 0) {$D^j_{f_o}$};
	\end{pgfonlayer}
	\begin{pgfonlayer}{edgelayer}
		\draw [style={arrow_label}] (1) to (0);
		\draw [style={arrow_label}] (2) to (0);
		\draw [style={arrow_label}] (0) to (11);
		\draw [style={arrow_label}] (9) to (8);
		\draw [style={arrow_label}] (7) to (10);
		\draw [dashed] (3) to (4);
		\draw [dashed] (3) to (5);
		\draw [dashed] (4) to (6);
		\draw [dashed] (5) to (6);
		\draw [dotted] (11) edge[->,below,red]  node        {$c_{kaf2f_{o}}$}   (7);
		\draw [dotted] (8) edge[->,above,red]  node        {$c_{f_{i}2plain}$}   (1);
		\draw [dotted] (8) edge[->,below,red]  node        {$c_{f_{i}2kaf}$}   (2);
	\end{pgfonlayer}
\end{tikzpicture}
    \caption{Wrapped $t_o$.}
    \label{fig:wrapto}
    \end{subfigure}\hfill{} 
    \begin{subfigure}[b]{0.5\textwidth}
 
\begin{tikzpicture}[scale=0.8, every node/.style={scale=0.8}]
	\begin{pgfonlayer}{nodelayer}
		\node [style=blackbox] (0) at (0, 0) {$t_{p}$};
		\node [style={empty_node}, label={above:[plain text]}] (1) at (-1, 0) {};
		\node [style={empty_node},label={above:[tabbed]}] (2) at (1, 0) {};
		\node [style=none] (3) at (-2.5, -1.5) {};
		\node [style=none] (4) at (-2.5, 1.5) {};
		\node [style=none] (5) at (2.5, -1.5) {};
		\node [style=none] (6) at (2.5, 1.5) {};
		\node [style=none] (7) at (2.5, 0) {};
		\node [style=none] (8) at (-2.5, 0) {};
		\node [style=none] (9) at (-3.5, 0) {$D^i_{f_i}$};
		\node [style=none] (10) at (3.5, 0) {$D^j_{f_o}$};
		
	\end{pgfonlayer}
	\begin{pgfonlayer}{edgelayer}
		\draw [style={arrow_label}] (1) to (0);
		\draw [style={arrow_label}] (0) to (2);
		\draw [style={arrow_label}] (9) to (8);
		\draw [style={arrow_label}] (7) to (10);
		\draw [dashed] (3) to (4);
		\draw [dashed] (3) to (5);
		\draw [dashed] (4) to (6);https://www.overleaf.com/project/5d9b04b9ce8bb6000190c19c
		\draw [dashed] (5) to (6);
		\draw [dotted] (8) edge[->,below,red]  node        {$c_{f_{i}2plain}$}   (1);
		\draw [dotted] (2) edge[->,below,red]  node        {$c_{tab2f_{o}}$}   (7);
	\end{pgfonlayer}
\end{tikzpicture}
     \caption{Wrapped $t_p$.}\label{fig:wraptp}

    \end{subfigure}
   \caption{Wrapped $t_o$ consume $f_i$ and produce $f_o$.}\label{fig:wrap}
\end{figure}
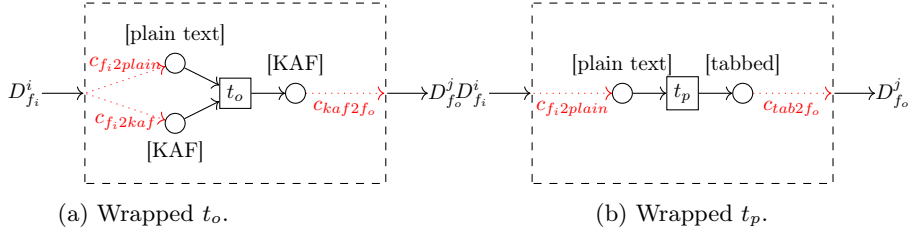

In this section, we describe the wrappers from the point of view of Category Theory. \\
According to our formalism, we can model the process as
\begin{equation}
    \label{eq:mainproc}
    D^i(c_i,f_i,\{a_i\})\xrightarrow{\tilde t_{ij}} D^j(c_j,f_j,\{a_j\})
\end{equation}

where we identify $W_{ij}$ with $\tilde t_{ij}$ since both format $f$, content $c$ and annotation set $\{a_j\}$ may change during the process, as actually they do. 
If we look at Figure \ref{fig:wrap} we see that the native tools $t_o$ and $t_p$, including the format converters, are wrapped into a wider box. In Figures \ref{fig:wrapto} and \ref{fig:wraptp}, the incoming format, $f_i$, takes values from $\{kaf,plain,tcf\}$ while the outgoing, $f_o$, from $\{kaf,tab,tcf\}$ 

\subsubsection{Morphisms}
\label{sssec:exmor}
The set of morphisms, i.e. the \acs{NLP} tools and converters between  documents $D$,  $Hom_{\mathcal{C}}(D^i,D^j)$, consists of both the original tools ($t_o,t_p$) and format converters. Such converters have the task of changing the input formats with those accepted by $t_o$ and $t_p$ as well as of  transforming the native output formats to one of $\{plain,tab,tcf\}$. To shorten the notation, we will agree on the following: a) $i$ runs on the set $\{kaf,plain,tcf\}$ while $j$ on $\{kaf,tab,tcf\}$; b) if $i=j$ the the input and the output formats of the converters are the same; c) $c_{i2j}$ stands from ``converting from format $i$ to format $j$''. According to points a), b) and c), input and output converters obey to the following rules:
\begin{equation}
\label{eq:iconvs1}
    c_{f_{i}2f_{j}} = \left\{
  \begin{array}{ll}
    c_{i2j}& : i \neq j\\
    id_{i} & : j = i\\
  \end{array}
\right.
\end{equation}
From Figure \ref{fig:wrap} we see that there are $12$ possible combinations and, thus, $12$ converters: $6$ of them manage incoming and $6$ outgoing formats. But when we consider the input and output restrictions of $t_o$ and $t_p$, we reduce the $12$ converters in definition \ref{eq:iconvs1} to $10$: $4$ converters for managing input and $6$ for output. We keep the $2$ identities in input and output,\footnote{Identities occur when incoming formats are either \textit{plain} or \textit{kaf}. In such cases, we don't need to convert such formats. The same happens when the native output of $t_o$ and $t_p$ are KAF and tabbed respectively.} and the necessary converters, see definitions \ref{eq:ci} and \ref{eq:co}.

\begin{subequations}
    \begin{align}\label{eq:ci}
      c^{input} = \left\{
  \begin{array}{rlllr}
    c_0& = id_{kaf} &;\text{ } 
    c_1& = id_{plain}\\
    c_2& = c_{tcf2kaf} &;\text{ }
    c_3& = c_{tcf2plain}\\
  \end{array}
\right.
    \end{align}
    \begin{align}\label{eq:co}
      c^{output} = \left\{
  \begin{array}{rlllr}
    c_4& = c_{kaf2tcf}&;\text{ }
    c_5& = c_{kaf2tab}\\
    c_6& = c_{tab2kaf}&;\text{ }
    c_7& = c_{tab2tcf}\\
    c_8& = id_{tab} &;\text{ }
    c_9& = id_{kaf}\\
  \end{array}
\right.
    \end{align}
    \end{subequations}
Please note that, thanks to our definition of $id$ as the ``do-nothing'' tool, $c_9$ in definition \ref{eq:co} is equivalent to $c_0$ in definition \ref{eq:ci}, so that the output converters reduce to $5$ ($9$ in total). We build the $Hom_{\mathcal{C}}(D^i,D^j)$ as in definition \ref{eq:homc}:
\begin{equation}\label{eq:homc}
Hom_{\mathcal{C}}(D^i,D^j)=\{t_o, t_p, id_{plain}, id_{kaf}, id_{tab},
 c_2,c_3,c_4,c_5,c_6,c_7\}
\end{equation}

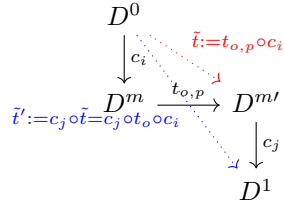
\begin{figure}[H]
\centering
\begin{tikzcd}
D^0 \arrow[d, "c_{i}"] \arrow[rd, "\tilde t := t_{o,p} \circ c_i", dotted,red] \arrow[rdd, "\tilde t^\prime :=c_j \circ \tilde t = c_j\circ t_o \circ c_i"', dotted,blue] &                               \\
D^m \arrow[r, "t_{o,p}"]                                                                                                                      & D^{m\prime} \arrow[d, "c_j"] \\
                                                                                                                                            & D^1                          
\end{tikzcd}
\caption{A diagram scheme for $D^0_{f_i} \xrightarrow{\tilde t_{01}} D^1_{f_o}$.}\label{fig:proc1}
\end{figure}

Finally, we have to remember that $\tilde t$ is either the composition of a converter $c$ and a tool $t$, $\tilde t \equiv c\circ t$, or the other way around a tool and a converter, $\tilde t \equiv t\circ c$. This ensure that, for some $c$ and $t$, $\tilde t$ belongs to $Hom_{\mathcal{C}}(D^i,D^j)$ as well. Each process in Figure \ref{fig:wrap} is identified by a diagram such the one in Figure \ref{fig:proc1}, where $c_i$ is one  of $\{c_2,c_3\}$ and $c_j$ one  of $\{c_4,c_5,c_6,c_7\}$.

Depending of $t$ being either $t_o$ or $t_p$, not all the compositions of $t$ with $c$ are possible. Indeed, $t_o$ is (input-)compatible with $\{c_2,c_3\}$ but only (output-)compatible with  $\{c_4,c_5\}$; while $t_p$ with $\{c_3\}$ and $\{c_6,c_7\}$ respectively. \\
Of course, if $c_i = id_i$ and $c_j = id_j$, the diagram in Figure \ref{fig:proc1} collapses to the one in Figure \ref{fig:proc2} where  $\tilde t$ is no longer needed.
\begin{figure}[H]
\centering
\begin{tikzcd}
D^0 \arrow[loop left]{r}{\mathrm{id}_{f_i}} \arrow[r, "t_{o,p}"] &
D^1 \arrow[loop right]{r}{\mathrm{id}_{f_o}} 
\end{tikzcd}
\caption{Simplified diagram scheme for $D^0_{f_i} \xrightarrow{t_{01}} D^1_{f_o}$.}\label{fig:proc2}
\end{figure}

\subsubsection{Objects}
\label{sssec:exobj}
We have to build the collection of objects, $Ob(\mathcal{C})$. In the process \ref{eq:mainproc}, $D^i$ is the initial  document: 
\begin{equation}
\label{eq:thetext01}
  D^0 = D(c,f_{i},\{\emptyset\})  \text{ with }\text{ a } \text {caveat}
\end{equation}

\begin{description}
\item[caveat:] in definition \ref{eq:thetext0} of the initial  document, $f_i$ is forced to be $f_p$ which corresponds to \textit{plain}. While here we assume that $f_{i}$ can be either \textit{kaf} or \textit{plain}. This should not surprise, since, on the one hand, it is related to the input restrictions of $t_o$ which accepts  either \textit{kaf} or \textit{plain} documents and, on the other hand, it is always possible to constrain $f_i$ to be $f_p$ by adding a converter $c_{kaf2plain}$. Such converter $c2=c_{tcf2plain}$ is in $Hom_{\mathcal{C}}(D^i,D^j)$. Therefore, we recover definition \ref{eq:thetext0} for the initial  document. Not to burden the category with (pretty much) useless morphisms we decided to release definition \ref{eq:thetext0} to \ref{eq:thetext01}.
\end{description}
where $c$ is the text to be analyzed, the incoming format $f_i$ is either \textit{plain}, \textit{kaf} or \textit{tcf}, and the annotation set in the empty set. We can apply converters $c_i$ to $D^0$ which leave content and annotation set unchanged: 
\[D^0_0 \xrightarrow{c_i} D^0_i\]
where, as usual, $c_i$ runs in $\{c_2,c_3\}$ and obtain $D^0_2$ and $D^0_3$. 
Then we can apply $t_o$ to the different $D^0_l$, with $l \in \{0,2,3\}$,  to add the annotation set obtaining $D^1_o$. Similarly, if we apply $t_p$ to $D^0_l$ we obtain $D^1_p$. Here we used the shorter notation: $D^1_o \equiv D^1_o(c,f_{out}=kaf,\{a_1\})$ and $D^1_p \equiv D^1_p(c^\prime,f_{out}o=tab,\{a_1\}$). Formats $f_{out}$ are tools' native formats, $\{a_i\}$ corresponds to the tokenization and $c^\prime$ in $D^i_p$ means that also the content $c$ is changed.

Finally, we can apply converters $c_j$ to $D^1_k$:
\[D^1_k \xrightarrow{c_j} D^1_{kj}\]
Since  $c_j$ runs in $\{c_4,c_5,c_6,c_7\}$ and $k$ runs in $\{o,p\}$, we obtain the following collections:
\[\{D^1_{o4},D^1_{o5},D^1_{o6},D^1_{o7},D^1_{p4},D^1_{p5},D^1_{p6},D^1_{p7}\}\]
We build the $Ob(\mathcal{C})$ as in definition \ref{eq:theobj}
\begin{multline}\label{eq:theobj}
Ob(\mathcal{C})=\{D^0,D^0_2,D^0_3 
D^1_{o4},D^1_{o5},D^1_{o6},D^1_{o7},D^1_{p4},D^1_{p5},D^1_{p6},D^1_{p7}\}
\end{multline}
\acresetall
\section{Future Work}
\label{sec:fw}
A possible research line is to use Category Theory to approach semantic interoperability. In this paper, we assumed that when a converter $c$ is applied to a document $D^i(c_i,f_i, \{a_i\})$ the resulting document is $D^j(c_i,f_j, \{a_i\})$, i.e. the annotation set $\{a_i\}$ is left unchanged. This is not generally true, because there are tools that need documents with different formats \textit{and} different values in $\{a_i\}$\footnote{This is the case, for example, when a tool $t$ needs `\texttt{VERB} instead of \texttt{V} as part-of-speech.} and we can't change $\{a_i\}$ with the $c$ and $t$ we have defined in $Hom_{\mathcal{C}}$. Thus, we have to improve the converters $c$ so that they can act on $\{a_i\}$:
$D^i(c_i,f_i, \{a_i\}) \xrightarrow{c} D^j(c_i,f_j, \{a_j\})$. Where the meaning of the values  in $\{a_j\}$ might differ from the one in $\{a_j\}$. Adding such converters to $Hom_{\mathcal{C}}$ allows us to define tools $\tilde t^\prime$ which perform the same linguistic operations as $\tilde t$ but work on a different set of values in $\{a_i\}$.
We can follow the strategy adopted by \cite{Cafezeiro:2007}  to model maps between the set of values in $\{a_i\}$ and $\{a_j\}$. In words, such morphisms are applications that maximally preserve the information when moving from $\{a_i\}$ to $\{a_j\}$. The question is are $\tilde t^\prime$ and $\tilde t$ \textit{``the same'' tool}? Do these \textit{maps} always exist? If so, are they unique? A different point of view is the following: if $D^i(c_i,f_i, \{a_i\}) \in Ob(\mathcal{C})$, is $D^j(c_i,f_j, \{a_i\}) \in Ob(\mathcal{C})$ as well? If we restrict the objects in $Ob(\mathcal{C})$ to have an annotation set $\{a\}$ with fixed values, the answer is negative. We can either relax this constraint or assume that $D^j(c_i,f_j, \{a_i\}) \in Ob(\mathcal{C^{\prime}})$  where $Ob(\mathcal{C^{\prime}})$ is a new category. Are $Ob(\mathcal{C})$ and $Ob(\mathcal{C^{\prime}})$ functorially connected?

\acresetall
\section{Conclusions}
We presented a Category Theory approach to syntactic interoperability. This approach allowed us to describe both the composition and associativity, typical issues of a chain of interoperable \acs{NLP} tools, through a more abstract mathematical formalism. The restrictions of input and output formats of the \acs{NLP} tools have been modeled as format converters.
The resulting category has the \acs{NLP} applications and the format converters as its morphisms, while the documents (with or without linguistic annotations) form its objects. We do not pretend to rewrite the approach to syntactic interoperability within the chains of linguistic tools, but we think that a more abstract approach to syntactic interoperability can help in the actual design and implementation of \acs{NLP} tools. Certainly, this approach captures the formal requirements of a tool in terms of its input/output specifications and of its linguistic operations providing a guide for software design and implementation. For example, looking at the  tools ($\tilde t$) as the composition of an \acs{NLP} tool and a format converter helps software engineers and programmers at keeping core tools (the tools the analyze documents) and format converters logically separated.\footnote{\ldots which,of course, does not means that, technically, tools and converters must be coded separately.} The Category Theory approach takes a step toward the implementation of atomic tools rather than complex ones, which is also in line with \cite{Dixit2017}, but given its abstraction, complex tools are also allowed. Or might be built. \\
 We also proposed  further investigations that involve more advanced concepts of Category Theory and that will be addressed in forthcoming papers. 

\bibliographystyle{abbrvnat}
\bibliography{ARXIV2020_Main}

\begin{thebibliography}{37}
\providecommand{\natexlab}[1]{#1}
\providecommand{\url}[1]{\texttt{#1}}
\expandafter\ifx\csname urlstyle\endcsname\relax
  \providecommand{\doi}[1]{doi: #1}\else
  \providecommand{\doi}{doi: \begingroup \urlstyle{rm}\Url}\fi

\bibitem[Aitor Garcia-Pablos(2013)]{OPENER}
G.~R. Aitor Garcia-Pablos, Montse~Cuadros.
\newblock {OpeNER demo: Open Polarity Enhanced Named Entity Recognition}.
\newblock pages 579--580, 2013.

\bibitem[Antunes and Abel(2018)]{AntunesA18}
C.~Antunes and M.~Abel.
\newblock Ontologies in category theory: {A} search for meaningful morphisms.
\newblock In J.~L. Carbonera and G.~Guizzardi, editors, \emph{Proceedings of
  the {XI} Seminar on Ontology Research in Brazil and {II} Doctoral and Masters
  Consortium on Ontologies, S{\~{a}}o Paulo, Brazil, October 1st-3rd, 2018.},
  volume 2228 of \emph{{CEUR} Workshop Proceedings}, pages 152--160.
  CEUR-WS.org, 2018.
\newblock URL \url{http://ceur-ws.org/Vol-2228/paper10.pdf}.

\bibitem[Awodey(2010)]{Awodey:2010:CT:2060081}
S.~Awodey.
\newblock \emph{Category Theory}.
\newblock Oxford University Press, Inc., New York, NY, USA, 2nd edition, 2010.
\newblock ISBN 0199237182, 9780199237180.

\bibitem[Baez and Stay(2011)]{Baez34405}
J.~Baez and M.~Stay.
\newblock {Physics, Topology, Logic and Computation: A Rosetta Stone}.
\newblock \emph{Lecture Notes in Physics}, 813:\penalty0 95--172, 2011.
\newblock URL \url{http://arxiv.org/abs/0903.0340}.

\bibitem[\bgroup Del Gratta~\egroup and Albanesi(2019)]{delgrattacac2019}
R.~\bgroup Del Gratta~\egroup and D.~Albanesi.
\newblock {OpeNER and PANACEA: Web Services for the CLARIN Research
  Infrastructure}.
\newblock In K.~Simov and M.~Eskevich, editors, \emph{Proceedings of CLARIN
  Annual Conference 2019, {CAC} 2019}, Leipzig, Germany, 2019.

\bibitem[\bgroup Del Gratta~\egroup et~al.(2016)\bgroup Del Gratta~\egroup,
  Boschetti, \bgroup Del Grosso~\egroup, Khan, and Monachini]{Springer2016}
R.~\bgroup Del Gratta~\egroup, F.~Boschetti, A.~\bgroup Del Grosso~\egroup,
  F.~Khan, and M.~Monachini.
\newblock \emph{{Cooperative Philology on the Way to Web Services: The Case of
  the CoPhiWordNet Platform}}, pages 173--187.
\newblock Springer International Publishing, Cham, 2016.
\newblock ISBN 978-3-319-31468-6.
\newblock \doi{978-3-319-31468-6_13}.

\bibitem[Boschetti et~al.(2017)Boschetti, \bgroup Del Gratta~\egroup, and
  \bgroup Del Grosso~\egroup]{Dixit2017}
F.~Boschetti, R.~\bgroup Del Gratta~\egroup, and A.~\bgroup Del Grosso~\egroup.
\newblock \emph{{The role of digital scholarly editors in the design of
  components for cooperative philology}}, pages 249--253.
\newblock 2017.
\newblock ISBN 978-90-8890-484-4.
\newblock URL
  \url{https://www.sidestone.com/books/advances-in-digital-scholarly-editing}.
\newblock ISBN 978-90-8890-483-7 (softcover) ISBN 978-90-8890-484-4 (hardcover)
  ISBN 978-90-8890-485-1 (PDF e-book).

\bibitem[Bosma et~al.(2009)Bosma, Vossen, Soroa, Rigau, Tesconi, Marchetti,
  Monachini, and Aliprandi]{KAF}
W.~Bosma, P.~Vossen, A.~Soroa, G.~Rigau, M.~Tesconi, A.~Marchetti,
  M.~Monachini, and C.~Aliprandi.
\newblock {KAF: a generic semantic annotation format}.
\newblock In \emph{Proceedings of the GL2009 Workshop on Semantic Annotation},
  2009.

\bibitem[Bradley(2018)]{2018arXiv180905923B}
T.-D. Bradley.
\newblock {What is Applied Category Theory?}
\newblock 2018.

\bibitem[Broeder et~al.(2012)Broeder, van Uytvanck, Gavrilidou, Trippel, and
  Windhouwer]{broeder-etal-2012-standardizing}
D.~Broeder, D.~van Uytvanck, M.~Gavrilidou, T.~Trippel, and M.~Windhouwer.
\newblock Standardizing a component metadata infrastructure.
\newblock In \emph{Proceedings of the Eighth International Conference on
  Language Resources and Evaluation ({LREC}'12)}, pages 1387--1390, Istanbul,
  Turkey, May 2012. European Language Resources Association (ELRA).
\newblock URL
  \url{http://www.lrec-conf.org/proceedings/lrec2012/pdf/581_Paper.pdf}.

\bibitem[Cafezeiro and Haeusler(2007)]{Cafezeiro:2007}
I.~Cafezeiro and E.~H. Haeusler.
\newblock {Semantic Interoperability via Category Theory}.
\newblock In \emph{Tutorials, Posters, Panels and Industrial Contributions at
  the 26th International Conference on Conceptual Modeling - Volume 83}, ER
  '07, pages 197--202, Darlinghurst, Australia, Australia, 2007. Australian
  Computer Society, Inc.
\newblock ISBN 978-1-920682-64-4.
\newblock URL \url{http://dl.acm.org/citation.cfm?id=1386957.1386989}.

\bibitem[Cieri et~al.(2010)Cieri, Choukri, Calzolari, Langendoen, Leveling,
  Palmer, Ide, and Pustejovsky]{CieriCCLLPIP10}
C.~Cieri, K.~Choukri, N.~Calzolari, D.~T. Langendoen, J.~Leveling, M.~Palmer,
  N.~Ide, and J.~Pustejovsky.
\newblock {A Road Map for Interoperable Language Resource Metadata}.
\newblock In N.~Calzolari, K.~Choukri, B.~Maegaard, J.~Mariani, J.~Odijk,
  S.~Piperidis, M.~Rosner, and D.~Tapias, editors, \emph{Proceedings of the
  International Conference on Language Resources and Evaluation, {LREC} 2010,
  17-23 May 2010, Valletta, Malta}. European Language Resources Association,
  2010.
\newblock ISBN 2-9517408-6-7.
\newblock URL
  \url{http://www.lrec-conf.org/proceedings/lrec2010/summaries/951.html}.

\bibitem[Coecke et~al.(2010)Coecke, Sadrzadeh, and
  Clark]{coecke2010mathematical}
B.~Coecke, M.~Sadrzadeh, and S.~Clark.
\newblock {Mathematical foundations for a compositional distributional model of
  meaning}.
\newblock \emph{arXiv preprint arXiv:1003.4394}, 2010.

\bibitem[Coecke et~al.(2013)Coecke, Grefenstette, and
  Sadrzadeh]{coecke2013lambek}
B.~Coecke, E.~Grefenstette, and M.~Sadrzadeh.
\newblock {Lambek vs. Lambek: Functorial vector space semantics and string
  diagrams for Lambek calculus}.
\newblock \emph{Annals of pure and applied logic}, 164\penalty0 (11):\penalty0
  1079--1100, 2013.

\bibitem[Cunningham(2000)]{Cun00a}
H.~Cunningham.
\newblock {Software Architecture for Language Engineering}, 2000.
\newblock URL \url{http://gate.ac.uk/sale/thesis/}.

\bibitem[Fellbaum(1998)]{WN}
C.~Fellbaum, editor.
\newblock \emph{WordNet: An Electronic Lexical Database (Language, Speech, and
  Communication)}.
\newblock The MIT Press, Cambridge, MA, USA, 1998.
\newblock ISBN 026206197X.
\newblock URL
  \url{http://www.amazon.ca/exec/obidos/redirect?tag=citeulike09-20\&amp;path=ASIN/026206197X}.

\bibitem[Ferrucci and Lally(2004)]{UIMA:FERRUCCI:2004}
D.~Ferrucci and A.~Lally.
\newblock {UIMA: An Architectural Approach to Unstructured Information
  Processing in the Corporate Research Environment}.
\newblock \emph{Natural Language Engineering}, 10\penalty0 (3-4):\penalty0
  327--348, sep 2004.
\newblock URL \url{https://doi.org/10.1017/S1351324904003523}.

\bibitem[Ferrucci et~al.(2009)Ferrucci, Lally, Verspoor, and
  Nyberg]{OASIS:UIMA:2009}
D.~Ferrucci, A.~Lally, K.~Verspoor, and E.~Nyberg.
\newblock {Unstructured Information Management Architecture ({UIMA}) Version
  1.0}.
\newblock OASIS Standard, mar 2009.
\newblock URL \url{https://docs.oasis-open.org/uima/v1.0/uima-v1.0.html}.

\bibitem[Francopoulo et~al.(2006)Francopoulo, Bel, George, Calzolari,
  Monachini, Pet, and Soria]{LMF}
G.~Francopoulo, N.~Bel, M.~George, N.~Calzolari, M.~Monachini, M.~Pet, and
  C.~Soria.
\newblock {Lexical Markup Framework (LMF) for NLP Multilingual Resources}.
\newblock In \emph{Proceedings of the Workshop on Multilingual Language
  Resources and Interoperability}, MLRI '06, pages 1--8, Stroudsburg, PA, USA,
  2006. Association for Computational Linguistics.
\newblock ISBN 1-932432-82-5.
\newblock URL \url{http://dl.acm.org/citation.cfm?id=1613162.1613163}.

\bibitem[Gavrilidou et~al.(2012)Gavrilidou, Labropoulou, Desipri, Piperidis,
  Papageorgiou, Monachini, Frontini, Declerck, Francopoulo, Arranz, and
  Mapelli]{gavrilidou-etal-2012-meta}
M.~Gavrilidou, P.~Labropoulou, E.~Desipri, S.~Piperidis, H.~Papageorgiou,
  M.~Monachini, F.~Frontini, T.~Declerck, G.~Francopoulo, V.~Arranz, and
  V.~Mapelli.
\newblock The {META}-{SHARE} metadata schema for the description of language
  resources.
\newblock In \emph{Proceedings of the Eighth International Conference on
  Language Resources and Evaluation ({LREC}'12)}, pages 1090--1097, Istanbul,
  Turkey, May 2012. European Language Resources Association (ELRA).
\newblock URL
  \url{http://www.lrec-conf.org/proceedings/lrec2012/pdf/998_Paper.pdf}.

\bibitem[Healy and Williamson(2000)]{healy2000}
M.~Healy and K.~Williamson.
\newblock {Applying Category Theory to Derive Engineering Software from Encoded
  Knowledge}.
\newblock In T.~Rus, editor, \emph{{Algebraic Methodology and Software
  Technology}}, pages 484--498, Berlin, Heidelberg, 2000. Springer Berlin
  Heidelberg.
\newblock ISBN 978-3-540-45499-1.
\newblock \doi{10.1007/3-540-45499-3_34}.
\newblock URL \url{https://doi.org/10.1007/3-540-45499-3_34}.

\bibitem[Hinrichs et~al.(2010)Hinrichs, Zastrow, and Hinrichs]{WL}
M.~Hinrichs, T.~Zastrow, and E.~Hinrichs.
\newblock {WebLicht: Web-based LRT Services in a Distributed eScience
  Infrastructure}.
\newblock In N.~C.~C. Chair), K.~Choukri, B.~Maegaard, J.~Mariani, J.~Odijk,
  S.~Piperidis, M.~Rosner, and D.~Tapias, editors, \emph{Proceedings of the
  Seventh International Conference on Language Resources and Evaluation
  (LREC'10)}, Valletta, Malta, may 2010. European Language Resources
  Association (ELRA).
\newblock ISBN 2-9517408-6-7.

\bibitem[Ide and Pustejovsky(2010)]{ide2010}
N.~Ide and J.~Pustejovsky.
\newblock {What does Interoperability Mean, Anyway? Toward an Operational
  Definition of Interoperability}.
\newblock In \emph{Proceedings of the Second International Conference on Global
  Interoperability for Language Resources ICGL 2010}, Hong Kong, China, 2010.

\bibitem[Ide and Suderman(2007)]{GRAF}
N.~Ide and K.~Suderman.
\newblock {GrAF: A Graph-based Format for Linguistic Annotations}.
\newblock In \emph{Proceedings of the Linguistic Annotation Workshop}, LAW '07,
  pages 1--8, Stroudsburg, PA, USA, 2007. Association for Computational
  Linguistics.
\newblock URL \url{http://dl.acm.org/citation.cfm?id=1642059.1642060}.

\bibitem[Ide et~al.(2009)Ide, Pustejovsky, Calzolari, and Soria]{IdePCS09}
N.~Ide, J.~Pustejovsky, N.~Calzolari, and C.~Soria.
\newblock The {SILT} and flarenet international collaboration for
  interoperability.
\newblock In \emph{Proceedings of the Third Linguistic Annotation Workshop,
  {LAW} 2009, August 6-7, 2009, Singapore}, pages 178--181. The Association for
  Computer Linguistics, 2009.
\newblock ISBN 978-1-932432-52-7.
\newblock URL \url{https://www.aclweb.org/anthology/W09-3034/}.

\bibitem[Ide et~al.(2014)Ide, Pustejovsky, Cieri, Nyberg, Wang, Suderman,
  Verhagen, and Wright]{ide-etal-2014-language-application}
N.~Ide, J.~Pustejovsky, C.~Cieri, E.~Nyberg, D.~Wang, K.~Suderman, M.~Verhagen,
  and J.~Wright.
\newblock The language application grid.
\newblock In \emph{Proceedings of the Ninth International Conference on
  Language Resources and Evaluation ({LREC}'14)}, Reykjavik, Iceland, May 2014.
  European Language Resources Association (ELRA).
\newblock URL
  \url{http://www.lrec-conf.org/proceedings/lrec2014/pdf/926_Paper.pdf}.

\bibitem[Ingersoll et~al.(2013)Ingersoll, Morton, and Farris]{TamingTxt}
G.~Ingersoll, T.~S. Morton, and A.~L. Farris.
\newblock Taming text: How to find, organize, and manipulate it.
\newblock 2013.

\bibitem[Lambek(2008)]{lambek2008word}
J.~Lambek.
\newblock \emph{{From Word to Sentence: a computational algebraic approach to
  grammar}}.
\newblock Polimetrica sas, 2008.
\newblock URL
  \url{http://www.math.mcgill.ca/barr/lambek/pdffiles/2008lambek.pdf}.

\bibitem[Loper and Bird(2002)]{NLTK}
E.~Loper and S.~Bird.
\newblock {NLTK: The Natural Language Toolkit}.
\newblock In \emph{Proceedings of the {ACL}-02 Workshop on Effective Tools and
  Methodologies for Teaching Natural Language Processing and Computational
  Linguistics}, pages 63--70, Philadelphia, Pennsylvania, USA, July 2002.
  Association for Computational Linguistics.
\newblock \doi{10.3115/1118108.1118117}.
\newblock URL \url{https://www.aclweb.org/anthology/W02-0109}.

\bibitem[Mac~Lane(1998)]{mclane1998}
S.~Mac~Lane.
\newblock \emph{{Categories for the Working Mathematician}}.
\newblock Graduate Texts in Mathematics. Springer, second edition, 1998.
\newblock ISBN 0387984038.
\newblock URL \url{http://www.worldcat.org/isbn/0387984038}.

\bibitem[Manning et~al.(2014)Manning, Surdeanu, Bauer, Finkel, Bethard, and
  McClosky]{manning-etal-2014-stanford}
C.~Manning, M.~Surdeanu, J.~Bauer, J.~Finkel, S.~Bethard, and D.~McClosky.
\newblock The {S}tanford {C}ore{NLP} natural language processing toolkit.
\newblock In \emph{Proceedings of 52nd Annual Meeting of the Association for
  Computational Linguistics: System Demonstrations}, pages 55--60, Baltimore,
  Maryland, June 2014. Association for Computational Linguistics.
\newblock \doi{10.3115/v1/P14-5010}.
\newblock URL \url{https://www.aclweb.org/anthology/P14-5010}.

\bibitem[Murakami et~al.(2010)Murakami, Lin, Tanaka, Nakaguchi, and
  Ishida]{murakami-etal-2010-language}
Y.~Murakami, D.~Lin, M.~Tanaka, T.~Nakaguchi, and T.~Ishida.
\newblock Language service management with the language grid.
\newblock In \emph{Proceedings of the Seventh International Conference on
  Language Resources and Evaluation ({LREC}'10)}, Valletta, Malta, May 2010.
  European Language Resources Association (ELRA).
\newblock URL
  \url{http://www.lrec-conf.org/proceedings/lrec2010/pdf/833_Paper.pdf}.

\bibitem[Odijk(2018)]{Odijk2018DiscoveringSR}
J.~Odijk.
\newblock {Discovering software resources in CLARIN}.
\newblock 2018.

\bibitem[Preller and Lambek(2007)]{preller_lambek_2007}
A.~Preller and J.~Lambek.
\newblock {Free compact 2-categories}.
\newblock \emph{Mathematical Structures in Computer Science}, 17\penalty0
  (2):\penalty0 309–340, 2007.
\newblock \doi{10.1017/S0960129506005901}.

\bibitem[Riehl(2017)]{riehl2017category}
E.~Riehl.
\newblock \emph{{Category Theory in Context}}.
\newblock Aurora: Dover Modern Math Originals. Dover Publications, 2017.
\newblock ISBN 9780486820804.
\newblock URL \url{https://books.google.it/books?id=6B9MDgAAQBAJ}.

\bibitem[Verhagen et~al.(2016)Verhagen, Suderman, Wang, Ide, Shi, Wright, and
  Pustejovsky]{LAPPSF}
M.~Verhagen, K.~Suderman, D.~Wang, N.~Ide, C.~Shi, J.~Wright, and
  J.~Pustejovsky.
\newblock {The LAPPS Interchange Format}.
\newblock In Y.~Murakami and D.~Lin, editors, \emph{Worldwide Language Service
  Infrastructure}, pages 33--47, Cham, 2016. Springer International Publishing.
\newblock \doi{10.1007/978-3-319-31468-6_3}.

\bibitem[Zinn(2018)]{Zinn2018}
C.~Zinn.
\newblock The language resource switchboard.
\newblock \emph{Comput. Linguist.}, 44\penalty0 (4):\penalty0 631--639, Dec.
  2018.
\newblock ISSN 0891-2017.
\newblock \doi{10.1162/coli_a_00329}.
\newblock URL \url{https://doi.org/10.1162/coli_a_00329}.

\end{thebibliography}
\onecolumn\newpage
\acresetall
\appendix

\section{\acl{LRT}}
\label{app:lrt}
A \ac{LR} is a machine-readable collection of data for written or spoken languages. A collection of texts of Homeric poems, an Italian dictionary, an English-Arabic (bilingual) dictionary, a specific edition of a book, a simple text are  \aclp{LR}. But  a list of words extracted from a book, the list of most frequent words used by Dante are \aclp{LR} as well. 

We often read sentences like ``the lexicon used by author X'' or ``this word is unusual for author Y'' when we go through some essays or criticisms, but also ``this concept is closer to the politician A than to B ''when we listen to public debates. The first pair of sentences is related to written data, while the second one to vocal data. Both of them, however, originate from information extracted from a \ac {LR}. Indeed, the lexicon of an author X is the list of distinct words used in the (literary) production of X and these words can be ranked according to their frequencies to obtain most and less frequent words. Or, a very deep analysis of speeches of politician A can extract opinions of A on some topics and so on.

We may ask how such information is extracted from \aclp{LR}. The answer is using \acl{LT}. \ac{LT} are the dynamic counterpart of \aclp{LR}. If the latter can be considered ``static'' in the sense that once created they are stable\footnote{This is not completely true. A \ac{LR} can be periodically updated, but between updates it is stable.}, the former perform linguistic tasks (in a given time span) to create or modify \aclp{LR} from data or an existent \ac{LR} respectively.
\[d \xrightarrow{LT} LR\]
\[LR \xrightarrow{LT_1} LR^{\prime}\]
Linguistic tasks may be complex, but the idea is simple. When, at school, in sentences such as ``Lysa likes oranges'', we assign the part of speeches (subject, verb, object\ldots) to words: \textit{Lysa} is a subject, \textit{likes} is a verb, \textit{oranges} is the object we are making \textit{part-of-speech tagging}. If we study the inter-dependency among words we are doing a \textit{parsing}. Or when we read an email and extract some information we are doing \textit{information extraction}. Things go more difficult when we try to understand the actual opinion of a person X on a topic Y or to classify some data according to a set of features. But, as humans, we are able to finish the tasks.

Language tasks can also be performed by machines. There is specially designed software to simulate the human ability to perform specific linguistic activities. Tools that process the natural language are part of the \ac{NLP} research field. 

There are many \ac{NLP} suites available. In addition to UIMA and GATE, we can cite CoreNLP\footnote{\url{https://stanfordnlp.github.io/CoreNLP/}} \cite{manning-etal-2014-stanford}, the Apache OpeNLP project\footnote{\url{https://opennlp.apache.org/}} \cite{TamingTxt} or the python-based NLTK\footnote{\url{https://www.nltk.org/}}, acronym which stands for Natural Language Tool Kit \cite{NLTK}.

On the site of \aclp{LR}, one of the most used, famous and powerful is WordNet\footnote{\url{https://wordnet.princeton.edu/}} \cite{WN}.  According to their website:
\begin{quote}
    WordNet\textregistered is a large lexical database of English. Nouns, verbs, adjectives and adverbs are grouped into sets of cognitive synonyms (synsets), each expressing a distinct concept. Synsets are interlinked by means of conceptual-semantic and lexical relations.
\end{quote}
\subsection{Metadating}
\label{subapp:md}
Metadata are data about data. In other words, metadata describe data. For example, a book tells a story. The story is the data contained in the book. However, the authors and the title bring additional information that is not necessary for the story told but might be useful for the book to be found using search engines. The same happens for \aclp{LR} and \acl{LT}. We can describe \ac{LRT} using metadata to say that ``A is a lexicon'', or that ``B is a parser'' and so on. 

Formally, metadata are pairs ``\textit{key}=\textit{value}'' whose meaning is described in a given schema\footnote{For example the Dublin Core\texttrademark \quad schemas at \url{https://www.dublincore.org/schemas/}}. The same \ac{LR}, however, can be described according to different schemas. This situation seems strange, but it's typical in the field of \ac{LRT}. 

Besides, metadata are not limited to describe what a \ac{LRT} is but they are also massively used to describe deep features of both \acp{LR} and \acp{LT}. For example, metadata are used to specify what an \acs{NLP} tool accepts as input and produces as output. Unfortunately, given the specificity of the field of \acp{LRT}, the possible values that \textit{value} can assume is an open set. In the case of part-of-speech tagger, a valid pair to specify the output is ``pos={V,N,A}''. But another part-of-speech tagger could use the alternative ``pos={VERB,NOUN,ADJECTIVE}''.
\section{Projects, \aclp{RI}, and Interoperability}
\label{app:ri}
From Appendix \ref{app:lrt} it seems that if we want to run an \acs{NLP} tool after another, we have only to use one of the available suites. Unfortunately, it is not so easy. And this happens for many reasons. Computational Linguistics, as a discipline, originates between the $1940s$ and $1950s$ in the United States as a mechanism to manage automatic translations. In Italy, Father R. Busa firstly applied computational methods to textual analysis. \acs{NLP} suites started to be available and robust $10$ to $20$ years ago. In the meantime, researchers in Computational Linguistics all over the world started to develop their proprietary software, using different methods, strategies, formats, and programming languages. When we come to use  \acs{NLP} suites, we find it quite difficult: the offered part-of-speech is \textit{not exactly} the one we are used to, and when we try to use our proprietary software through such suites, well our tools often are not compliant with the suite specifications, precisely for interoperability reasons. Also, what if we have a lot of data in our data centers we can't run \acs{NLP} tools on? 

It is needed, then, to expand the concepts of \acs{NLP} suites and data centers. Here is where projects and \aclp{RI} come to play.

Platforms such as The Language Application Grid, lapps, \cite{ide-etal-2014-language-application}, \url{https://www.lappsgrid.org/}, in the US, the Language Grid \cite{murakami-etal-2010-language}, \url{https://langrid.org/en/index.html}, in Japan, European Projects such as PANACEA,  \url{http://www.panacea-lr.eu/}, and OpeNer \cite{OPENER}, \url{http://www.opener-project.eu/} are \textit{an evolution} of \acs{NLP} suites. Lapps fosters interoperability \cite{LAPPSF}; the same holds true for OpeNer and PANACEA. Indeed, we see the adoption of \ac{KAF} \cite{KAF} in OpeNer,\footnote{\url{https://github.com/opener-project/kaf/wiki/KAF-structure-overview}} of \ac{GrAF} \cite{GRAF} in PANACEA\footnote{\url{http://www.panacea-lr.eu/system/graf/graf-TO2_documentation_v1.pdf}} and  \ac{LMF} \cite{LMF} in both projects as a clear direction towards interoperability. 

But it is with \aclp{RI} that many research communities made further steps. 
\aclp{RI}\footnote{\url{https://ec.europa.eu/info/research-and-innovation/strategy/european-research-infrastructures\_en}}
\begin{quote}
    are facilities that provide resources and services for research communities to conduct research and foster innovation.
\end{quote}
There are \acp{RI} for public service, (high-energy) physics, health \ldots And for Computational Linguistics and the sub-field of \ac{SSH}.  but entities\footnote{They are ERIC, which stands for European Research Infrastructure Consortia.} such as CLARIN, \url{https://www.clarin.eu},  and DARIAH, \url{https://www.dariah.eu/} are proper \aclp{RI} that ``provide resources and services \ldots to conduct research \ldots''. According to CLARIN manifesto,
\begin{quote}
    [CLARIN] makes digital language resources available to scholars, researchers, students and citizen-scientists from all disciplines, especially in the \acl{SSH}.
\end{quote}
while DARIAH's states:
\begin{quote}
    The Digital Research Infrastructure for the Arts and Humanities (DARIAH) aims to enhance and support digitally-enabled research and teaching across the arts and humanities\ldots
\end{quote}
Both of them foster interoperability, of course. For example, in CLARIN, WebLicht \cite{WL} and \ac{LRS} \cite{Zinn2018} offer linguistic chains based on an agreed structure, the T\"ubingen Corpus Format, TCF\footnote{The TCF format is described at \url{https://weblicht.sfs.uni-tuebingen.de/weblichtwiki/index.php/The_TCF_Format}.}, along with a specific metadata format, the CMDI\footnote{\url{https://www.clarin.eu/content/component-metadata}.} \cite{broeder-etal-2012-standardizing},  and a metadata description (in JSON) that provides the relevant information for executing the tools.

\subsection{Metadating and Interoperability}
\label{subapp:mdi}
In Appendix \ref{subapp:md}, we enumerated two cases when the same \ac{LR} or Technology is described with two different metadata schemas and when the same pair ``\textit{key}=\textit{value}'' is applied, but the field \textit{value} is different.

The former happened, for example, with Metashare\footnote{\url{http://www.meta-share.org/}} and CLARIN\footnote{Before being an ERIC, CLARIN was a European project. Project in which the technological bases of the future ERIC have been defined.}. Metashare is slightly later than CLARIN, but decided to implement its own metadata schema \cite{gavrilidou-etal-2012-meta} rather than use CLARIN's CMDI schema. Then, if the same resource $R$ is described according to the two schemas, there should be a \textit{syntactic} mapping\footnote{This does not occur, of course. Indeed, a bijection between two schemas seldom exists.} from one to another:
\[R_{ms} \xleftrightarrow{}R_{clarin}\]

The latter is related to \textit{semantic} interoperability. Given the same schema, a \textit{semantic} mapping\footnote{As for syntactic interoperability, a complete semantic mapping is far from being reached. In Computational Linguistics, there are cases when a value, for instance, \texttt{VERB}, is mapped from two (or even more) different values, for instance, a transitive and an intransitive verb, \texttt{VI}, \text{VT}. It is always possible to map from fine to a coarse-grained value, but the vice-versa can not be done. What we can say is that one of the possible (fine-grained) value belongs to the preimage of \texttt{VERB}.  } from one set to another:
\[\{V, N, A\} \xleftrightarrow{}\{VERB, NOUN, ADJECTIVE\}\]

\section{Linguistic Annotations}
\label{app:la}
Linguistic annotation is additional information someway attached to a text, a part of the text, a single word, or a single character. Without pretending to be linguistically rigorous, we provide an example\footnote{Usually, annotations obey to a schema} Given the sentence ``Lysa likes oranges'', a human or a machine can annotate it as follows:
\begin{figure}[H]
    \centering
    
\begin{verbatim}
   a) <SENTENCE>Lysa likes oranges</SENTENCE>

   b) <SUBJ>Lysa</SUBJ> <VERB>likes</VERB> <OBJ>oranges</OBJ>

   c) <CAP>L</CAP>ysa 
      <VERB type="3rd singular person" verb="like">likes</VERB> 
      <NAME type="plural" name="orange">oranges</NAME>
\end{verbatim}
\caption{Some examples of inline tags for linguistic annotations.}
    \label{fig:lainline}
\end{figure}
Many other annotations are possible.
Annotations in Figure \ref{fig:lainline} are called \textit{inline}, because the tags \texttt{<../>} they use  are directly inserted  in text. If a person reads the annotations, [s]he gets from a) that ``Lysa likes oranges'' is something called \texttt{SENTENCE}; from b) that Lysa has the role of \texttt{SUBJ}; from c) that L in Lysa is a capital letter \texttt{CAP} \ldots

Different annotations provide different information. A person, a human agent, can understand the meaning of the various tags: SENTENCE, SUBJ \ldots However, a machine-based agent, an \acs{NLP}, can be told how to deal with such tags.

In addition to the inline annotations, there are the standoff ones. Standoff annotation means that all tags are moved from the text which is left unchanged.

\begin{figure}[H]
    \centering
    
\begin{verbatim}
   <TEXT>Lysa likes oranges</TEXT>
   ...
   <word id=1>Lysa</word>
   <word id=2>likes</word>
   <word id=3>oranges</word>
   ...
   <ROLES>
        <ROLE wid=1 type="SUBJ"/>
        <ROLE wid=2 type="VERB"/>
        <ROLE wid=1 type="OBJ"/>
   </ROLES>
\end{verbatim}
\caption{An example of standoff annotation.}
    \label{fig:lastandoff}
\end{figure}
The annotation in Figure \ref{fig:lastandoff} replaces annotation b) in Figure \ref{fig:lainline}. It provides the same information as b) does: Lysa has the role of \texttt{SUBJ}, likes of \texttt{VERB} and so on but using a different format. The original sentence is split by tokens; to each token is assigned a n identifier and additional information is connected to the identifier. 

The above mentioned \acs{KAF} and \acs{GrAF} are standoff annotation schemas, while, for example, the \ac{TEI}, \url{https://tei-c.org/}, is inline.
\end{document}